\pdfoutput=1
\documentclass{article}

\usepackage{microtype}
\usepackage{graphicx}
\usepackage{subfigure}
\usepackage{booktabs} % for professional tables
\usepackage{comment}
\usepackage{graphicx}
\usepackage{hyperref}
\usepackage[accepted]{icml2019}
 
\usepackage{icml2019}

\icmltitlerunning{Few-Shot Point Cloud Region Annotation with Human in the Loop}

\begin{document}

\twocolumn[
\icmltitle{Few-Shot Point Cloud Region Annotation with Human in the Loop}



\icmlsetsymbol{equal}{*}

\begin{icmlauthorlist}
\icmlauthor{Siddhant Jain\textsuperscript{*$\dagger$}}{cmu}
\icmlauthor{Sowmya Munukutla\textsuperscript{*$\dagger$}}{cmu}
\icmlauthor{David Held}{cmu}
\end{icmlauthorlist}

\icmlaffiliation{cmu}{Robotics Institute, Carnegie Mellon University, Pittsburgh, PA, USA}

\icmlcorrespondingauthor{Siddhant Jain}{siddhanj@andrew.cmu.edu}
\icmlcorrespondingauthor{Sowmya Munukutla}{spmunuku@andrew.cmu.edu}
\icmlcorrespondingauthor{David Held}{dheld@andrew.cmu.edu}

\icmlkeywords{Few shot learning, Point cloud segmentation, interactive learning, active learning design, annotation framework }

\vskip 0.3in
]



\printAffiliationsAndNotice{\icmlEqualContribution, \icmlNiantic} % otherwise use the  standard text.
%\printAffiliationsAndNotice{\icmlNiantic} % otherwise use the standard text.

\begin{abstract}
We propose a point cloud annotation framework that employs human-in-loop learning to enable the creation of large point cloud datasets with per-point annotations. Sparse labels from a human annotator are iteratively propagated to generate a full segmentation of the network by fine-tuning a pre-trained model of an allied task via a few-shot learning paradigm. We show that the proposed framework significantly reduces the amount of human interaction needed in annotating point clouds, without sacrificing on the quality of the annotations. Our experiments also suggest the suitability of the framework in annotating large datasets by noting a reduction in human interaction as the number of full annotations completed by the system increases. Finally, we demonstrate the flexibility of the framework to support multiple different annotations of the same point cloud enabling the creation of datasets with different granularities of annotation. 

Code and additional results are available  \href{https://github.com/siddhantjain/PointCloudAnnotationTool}{here}.

\end{abstract}

\section{Introduction}
\label{introduction}
Two dimensional images have been the most popular digital representation of the world however, point cloud data is increasingly gaining center stage with applications in autonomous driving, robotics and augmented reality. While synthetic point cloud datasets have been around for some time \cite{shapenet2015}, prevalance of depth cameras such as \cite{8014901} and \cite{Zhang:2012:MKS:2225053.2225203} has led to creation of large 3D datasets \cite{Vladlen} created from applying techniques from \cite{Newcombe} on depth scans. Finally, we have also seen a number of point cloud datasets created using LIDAR scans of outdoor environments such as \cite{hackel2017isprs}, \cite{Behley2019ADF}. 

The intensity and geometric information in point clouds provide a more detailed digital description of the world than images but their value in algorithmic analysis is fully realised when the points have an associated semantic label. However, annotating 3D point clouds is a time-consuming and labour intensive process owing to the size of the datasets and the limitations of the 2D interfaces in manipulating 3D data. 

The problem of providing a label of each point in a point cloud has been tackled via a host of fully automatic approaches in the domain of point cloud segmentation \cite{DBLP:journals/corr/abs-1710-07563} \cite{DBLP:journals/corr/abs-1711-09869}. While these approaches are successful in delineating large structures such as buildings, roads and vehicles, they perform poorly on finer details in the 3D models. Besides, most of these approaches use supervised learning methods which in-turn rely on labelled datasets making it a chicken and egg problem. 

Thus, most existing datasets \cite{Behley2019ADF} \cite{hackel2017isprs} \cite{Roynard2018ParisLille3DAL} have been annotated via dominantly manual systems to ensure accuracy and to avoid algorithmic biases in the produced datasets. The large investment required in terms of human effort in generating the annotations severely limit both the significance and prevalence of point cloud datasets that are available for the community.

Annotating large scale datasets is a natural use case for fusing human and algorithmic intelligence. Annotations inherently rely on a human definition and are also representative of semantic patterns that can be identified by an algorithm. Thus, we observe an active field of research which seeks to fuse human and algorithmic actors in one overarching framework to aid in annotating datasets. Most notable in the context of point cloud annotation is \cite{Yi:2016:SAF:2980179.2980238} which proposes an active learning framework for region annotations in datasets with repeating shapes. However, this method is limited by allowing for annotating only in certain 2D views the point clouds. Our proposed framework allows for annotation in full 3D thus allowing for finer annotation of the point cloud and the ability to work with less structured real-world point clouds as opposed to relatively noise-free synthetic point clouds.           
\begin{figure*}[h]
        \centering \includegraphics[width=\textwidth]{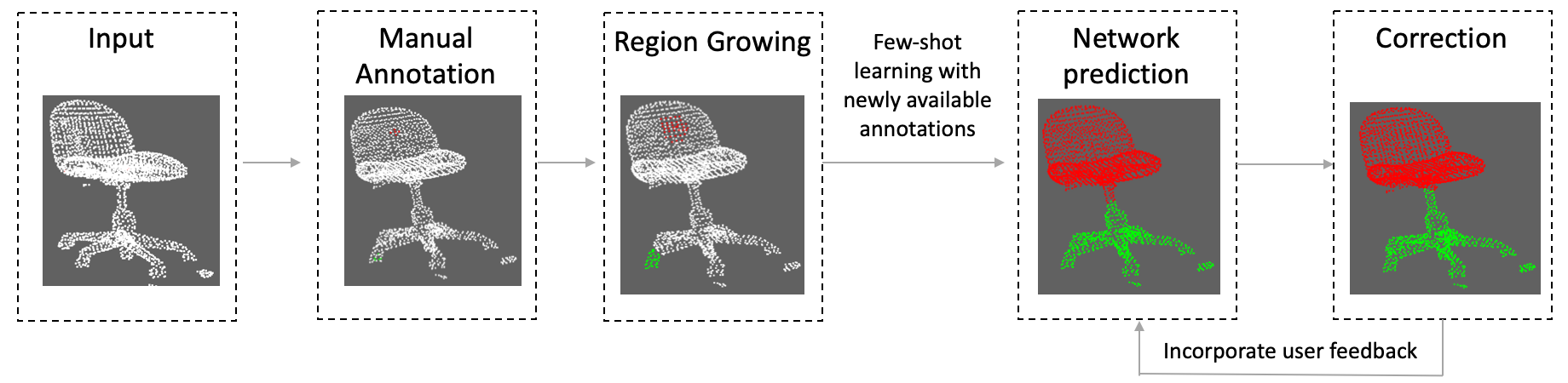}
        \caption{
                \label{fig:pipeline} 
                This figure illustrates our pipeline. The first step starts with a partial sparse annotation by a human, followed by a region growing step using 3D geometric cues. We then iterate between few-shot learning using newly available annotations and sparse correction of predictions via human annotator to obtain final segmentation outputs.
        }
\end{figure*}

In this work, we propose a human-in-loop learning approach that fuses together manual annotation, algorithmic propagation and capitalises on existing 3D datasets for improving semantic understanding. Our method starts with a partial sparse annotation by a human, followed by a region growing step using 3D geometric cues. We  then iterate over the following steps: a) Model fine tuning using newly available annotations b) Model prediction of annotated point cloud c) Sparse correction of predictions via human annotator. Figure \ref{fig:pipeline} gives a snapshot of our annotation approach. 

In the next couple of sections we go over our methodology in more details followed by a discussion of our results and future work.

\section{Methodology}
\label{method}
This system is primarily focused on providing an annotation framework to create datasets of point clouds with ground truth semantic labels for each point. For a given point cloud, our method starts with sparse manual annotation and then iterates between two main steps: few-shot learning and manual correction. The manual annotations are provided by marking few representative points for each part to be labelled in the point cloud. These labels are propagated across the point cloud using geometric cues which is used to train the network. The final step involves correcting network mispredictions, which is used to further guide the training process. For the initial point clouds to be annotated, these steps are iterated over multiple times but as more point clouds are annotated using this framework, the method converges to relying only on the initial set of manual annotations (or no annotations at all) to make more accurate annotation predictions.

\subsection{Manual Annotations and Region Growing}
The decomposition of a point cloud into semantically meaningful parts/regions is an open-ended problem as the concept of an annotation is context dependent \cite{segmentation_eval}. Owing to this ambiguity, the first step of our annotation pipeline is to allow the user to determine the number of possible classes that exist in the segmentation of point clouds in the dataset. The framework of annotation, learning and correction also provides the flexibility to have different number of segmentations for the same point cloud allowing for creating datasets with varying granularities of segmentation as in \cite{DBLP:journals/corr/abs-1812-02713}. The user initially provides labels to a point  or a small group of points for each of the classes in the point cloud. Thereafter, human provided annotations are automatically propagated to few unlabelled points by exploiting geometry of the point cloud. We believe that relying on geometric attributes like surface normals, smoothness, curvature and color (if available) would simplify the goal of segmentation as decomposing the point cloud into locally smooth regions enclosed by sharp boundaries. These segmentations also often end up matching with human perception and can be used as an initial training example for the learning pipeline. For this reason, we use cues like surface normals to group spatially close points as belonging to the same region. We also experimented with color based region growing, K-Nearest Neighbour (KNN) and Fixed Distance Neighbor (FDN) \cite{nearest_neighbor} based region growing methods which end up being faster than surface normal based region growing methods without compromising on accuracy of the overall system. Region growing approaches reduce annotator overhead of selecting multiple points by giving a geometry aware selection mechanism. 

\subsection{Few-shot Learning}
The goal of few-shot learning optimization in this context is to rely on minimal human supervision to improve segmentation accuracy. It is for this reason that we obtain the initial set of ground truth labels for training from manual annotations and use region growing methods for further supervision. We use very conservative thresholds for the region growing methods to avoid noisy ground truth labels. We also use a pre-trained network to bootstrap the training process and reduce the amount of human effort in correction phase.

The pre-trained network to be used in this system can be any segmentation network, pre-trained on an existing dataset in a similar domain. For our experiments, we used PointNet \cite{pointnet} pre-trained on ShapeNet \cite{DBLP:journals/corr/ChangFGHHLSSSSX15} to bootstrap the training.

\begin{figure}[h] 
        \centering \includegraphics[width=0.9\columnwidth]{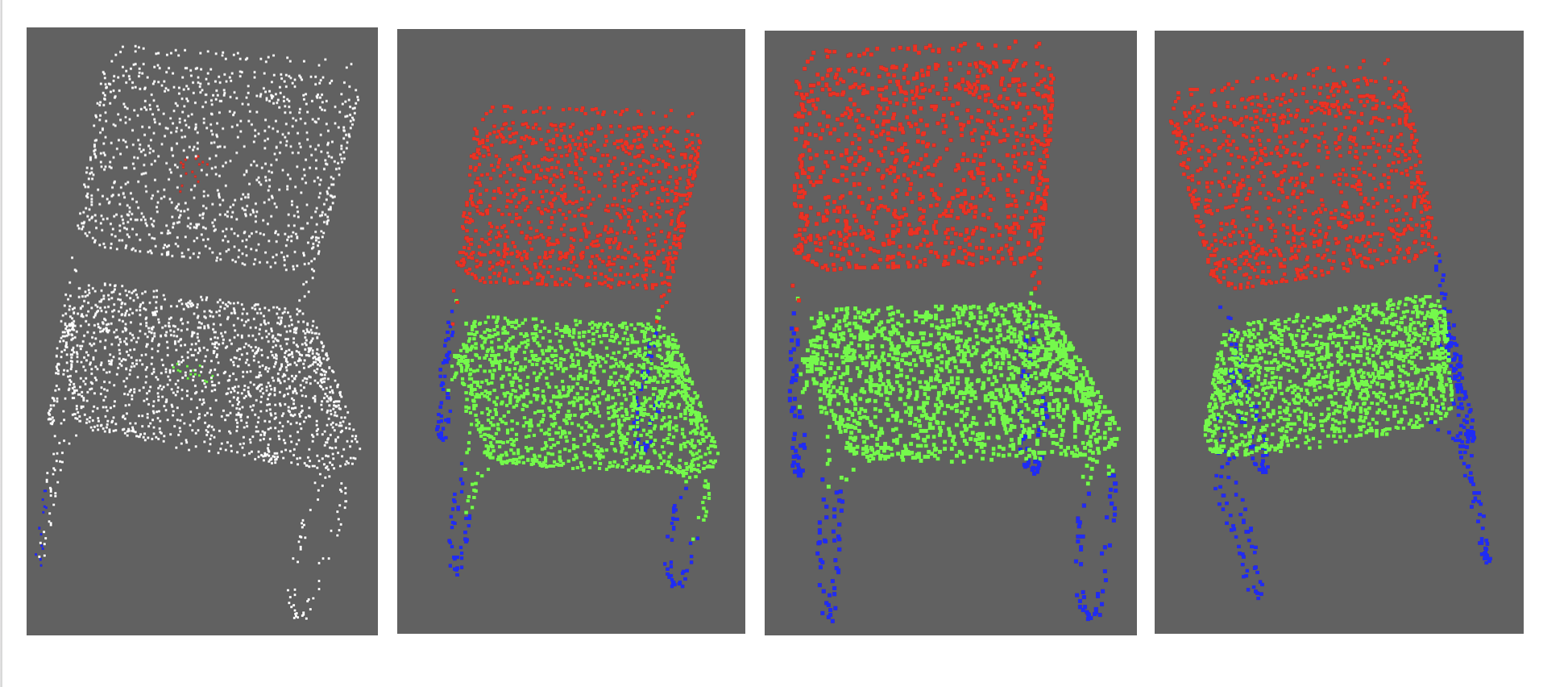}
        \caption{
                \label{fig:FineTuningExample} 
                Figure to show effects of few-shot learning in 3 class segmentation of a chair. From left to right i) Manual annotation with region growing ii) Predictions of the network after fine tuning. Notice the spillage of labels at the boundary which is resolved after correction and final learning step  iii) Partially corrected point cloud from the user iv) Final prediction after fine tuning with corrections
        }
\end{figure}

We fine-tune the base network iteratively using limited supervision in the form of annotation and correction provided by human in the loop. We also dynamically adapt the base network depending on the number of segmentation classes in the point cloud. The initial seed acquired from manual annotation and region growing gives a partially labelled point cloud that is used to fine-tune the base network. The model leverages the prior semantic understanding in the pre-trained network alongside the supervision of partially labelled points in the entire point cloud to provide meaningful segmentations in the first stage. We rely on the human annotators to compensate for network mispredictions by assigning new label to points with incorrect segmentations. Subsequently, we fine-tune further with all the labels (initial seed + corrections) that the human annotator has provided so far. This process continues until all points are labelled correctly - as verified by the human annotator. At this stage, we retrain the network with all the points in the point cloud - which allows us propagate these labels to newer point clouds of the same shape in the dataset. Figure \ref{fig:FineTuningExample} illustrates a sample of the results from this loop of user feedback and finetuning.   

% I think this is a very trivial detail to mention
%Based on the number of classes in a point cloud (as configured by the user), we change the last fully connected layer to allow for different number of classes in different annotation use-cases. 

\subsubsection{Smoothness Loss}
We formulate segmentation as a per-point classification problem similar to the setup of PointNet \cite{pointnet} including global and local feature aggregation. We also use transformation network to ensure that the network predictions are agnostic to rigid transformations of point cloud. We further leverage smoothness of the shape to favor regions that are compact and continuous. Overall, the network loss can be formulated as:
\begin{equation}
\mathcal{L} = L_{segment} + \alpha L_{transform} + \beta L_{smooth}
\end{equation}

We use \textit{smoothness loss} in addition to the segmentation cross entropy loss to encourage adjacent points to have similar labels. The smoothness loss is formulated as follows: \\
\begin{equation}
L_{smooth} = \sum_{i=1}^{N}\sum_{j=i}^{N}D_{KL} \left( p_i \middle\| p_j \right) e^{-\frac{\sqrt{\left\Vert (pos_{p_i} - pos_{p_j})\right\Vert_2^2}}{\sigma}}
\end{equation}

The smoothness term is computed as pairwise \textit{Kullback-Leibler} (KL) divergence of predictions exponentially weighted on eucledean distance between any two points in the point cloud. $\sigma$ is set to the variance of pairwise distance between all points to capture point cloud density in the loss term. The smoothness term in this context is expected to capture and minimize relative entropy between neighboring points in the point cloud. This term ends up dominating total loss if nearby points have divergent logits. Points which are far from each other end up contributing very little to the overall loss term, regardless of their logits owing to high pairwise distances between them. 

\begin{figure}[h] 
        \centering \includegraphics[width=0.8\columnwidth]{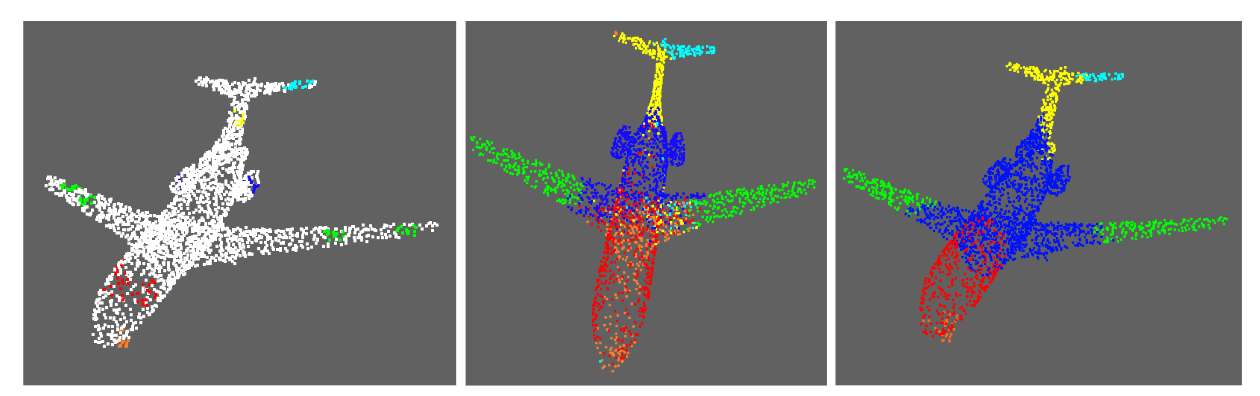}
        \caption{
                \label{fig:SmoothnessLossExample} % 
                Illustration to show effect of the smoothness loss. From left to right i) Manual annotation with region growing ii) Predictions of the network without smoothness loss iii) Network predictions with smoothness loss
        }
\end{figure}

Figure \ref{fig:SmoothnessLossExample} shows a qualitative example for the effect of the smoothness loss.

The first stage of segmentation output requires less human cognitive effort for correction if the smoothening term is added to loss computation as it has been observed through our experiments. The weights of smoothness loss term is subsequently dropped after getting further supervision from the user.

\section{Results}
\label{results}
In this section, we discuss the experimental setup to validate the effectiveness of our framework by investigating its utility to create new datasets against completely manual or semi-automatic methods. Additionally, we have also investigated the improvement in annotation efficiency as the total number of annotated point clouds increase. 

\begin{figure}[h] 
        \centering \includegraphics[width=0.8\columnwidth]{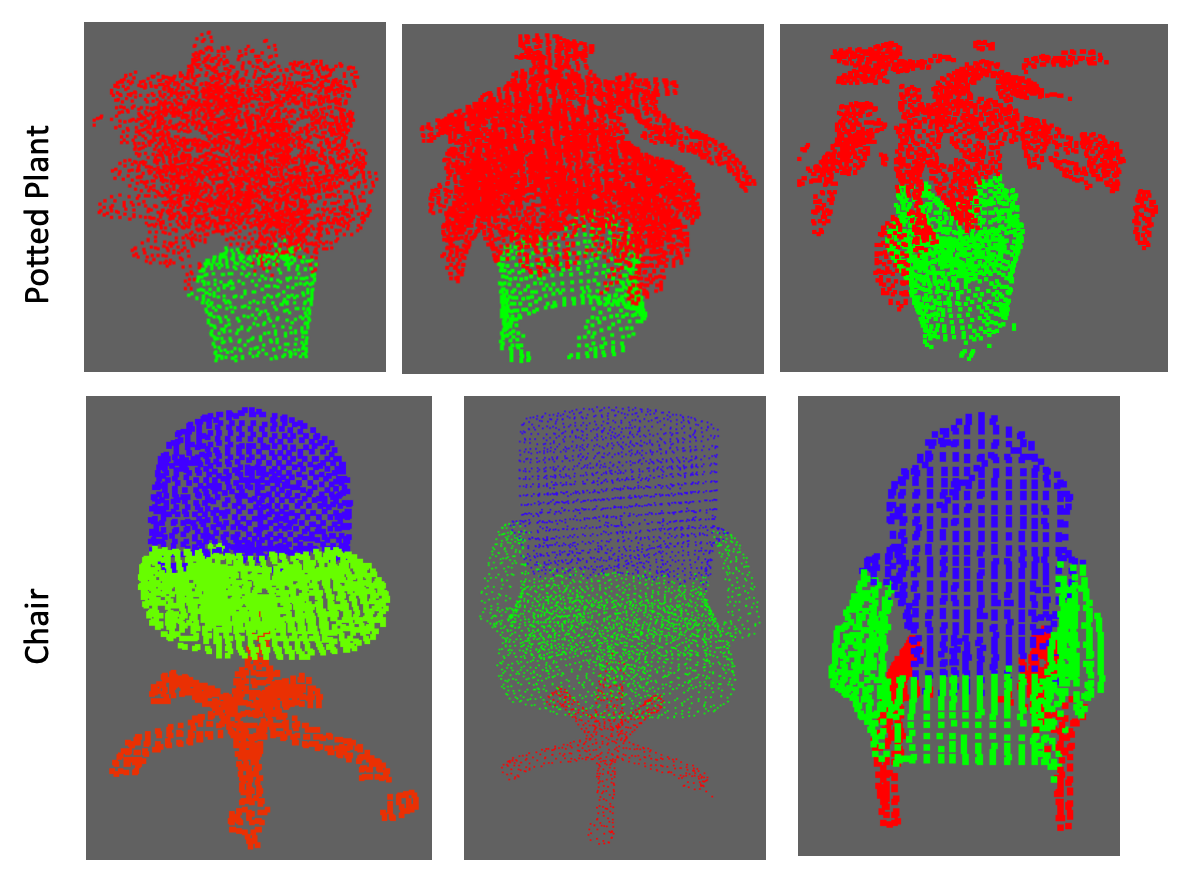}
        \caption{
                \label{fig:VladenDataset} % 
                Qualitative results for part segmentation on reconstructed point clouds from Large Dataset of Object Scans \cite{Vladlen}. The results are shown for segmentation of noisy shapes in potted plant and chair class into two and three classes respectively using our framework.
        }
\end{figure}

\textbf{Dataset}. To test the robustness and ease of adapting to our framework, we aim to use it to create a massive and diverse dataset of synthetic and reconstructed point clouds. Towards this goal, we have created part segmentations of reconstructed point clouds taken from A Large Dataset of Object Scans \cite{Vladlen}. Qualitative results for segmentations are shown in Figure \ref{fig:VladenDataset}. The framework showed remarkable improvement in human annotation efficiency as measured in number of clicks required for manual annotation and correction which is discussed in subsequent parts of this section. 

\begin{figure}[h] 
        \centering \includegraphics[width=0.8\columnwidth]{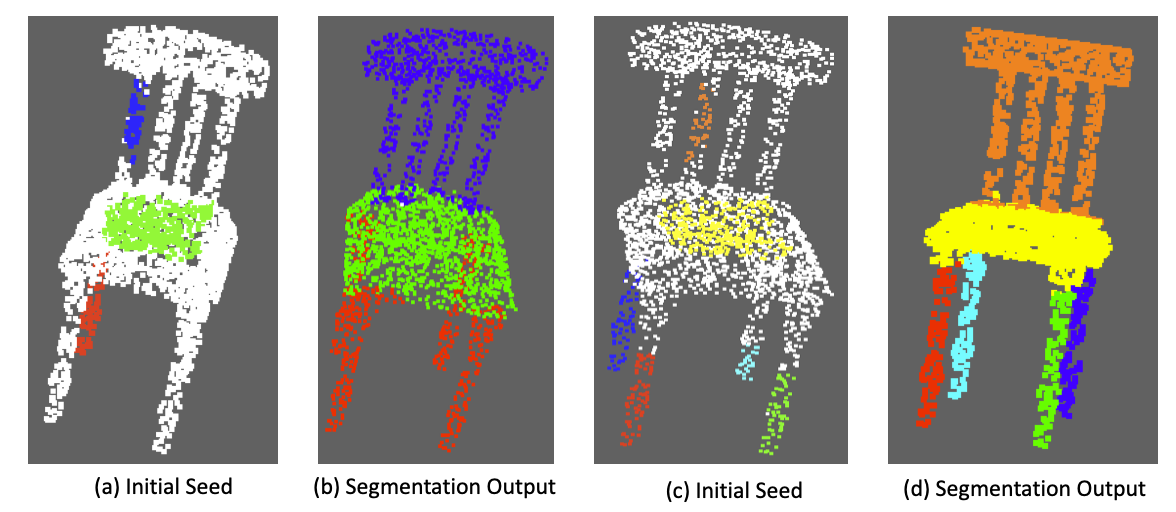}
        \caption{
                \label{fig:chair} % 
                Qualitative results for part segmentation on the same shape with different granularities.
        }
\end{figure}

\textbf{Granularity}. The framework also provides the flexibility to annotate with different number of classes for the same shape. The user selects sparse points for each of the classes in the first stage and this information is dynamically incorporated in the training process by re-initializing the last layer to accommodate different number of classes. Qualitative segmentation outputs are illustrated in Figure \ref{fig:chair}.

\subsection{Annotation Efficiency Improvements}
Existing semi-supervised methods \cite{Yi:2016:SAF:2980179.2980238} use amount of supervision and accuracy as evaluation metrics to measure performance. We follow suit and compare the amount of supervision needed to completely annotate a point cloud in our framework as opposed to completely manual methods. In Table \ref{table1} we compare the number of clicks by the annotator required in our framework compared to a naive nearest neighbour painting based manual approach.  

\begin{table}
  \caption{Average number of clicks taken to annotate point clouds with varying granularities in terms of number of parts for the same shape. We notice a significant reduction in number of clicks in comparison to manual methods and even our method without the smoothness constraint.}
  \label{table1}
  \includegraphics[width=\linewidth]{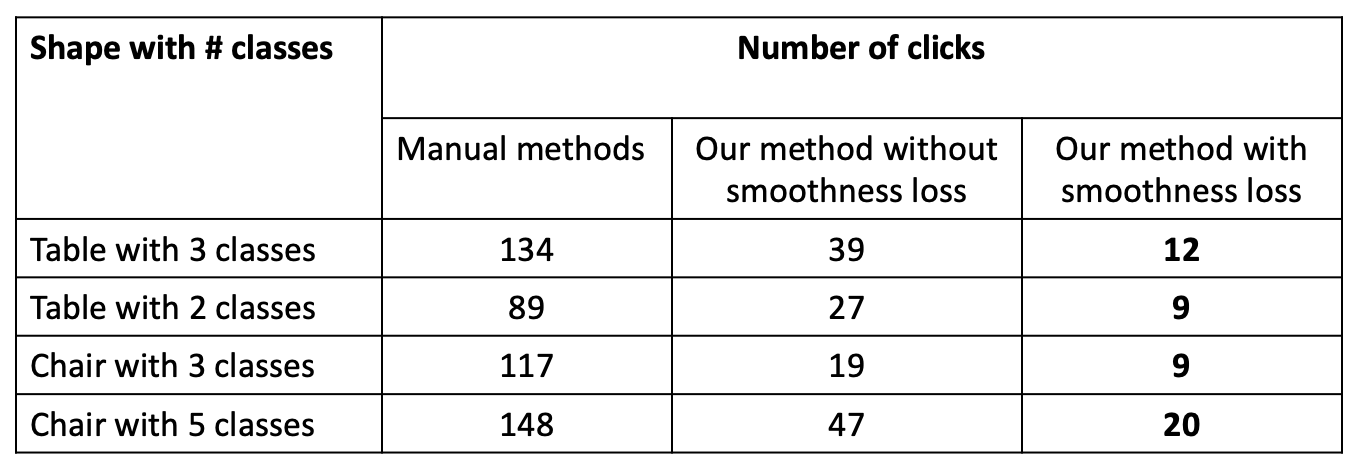}
\end{table}

\begin{figure}[h] 
        \centering \includegraphics[width=0.8\columnwidth]{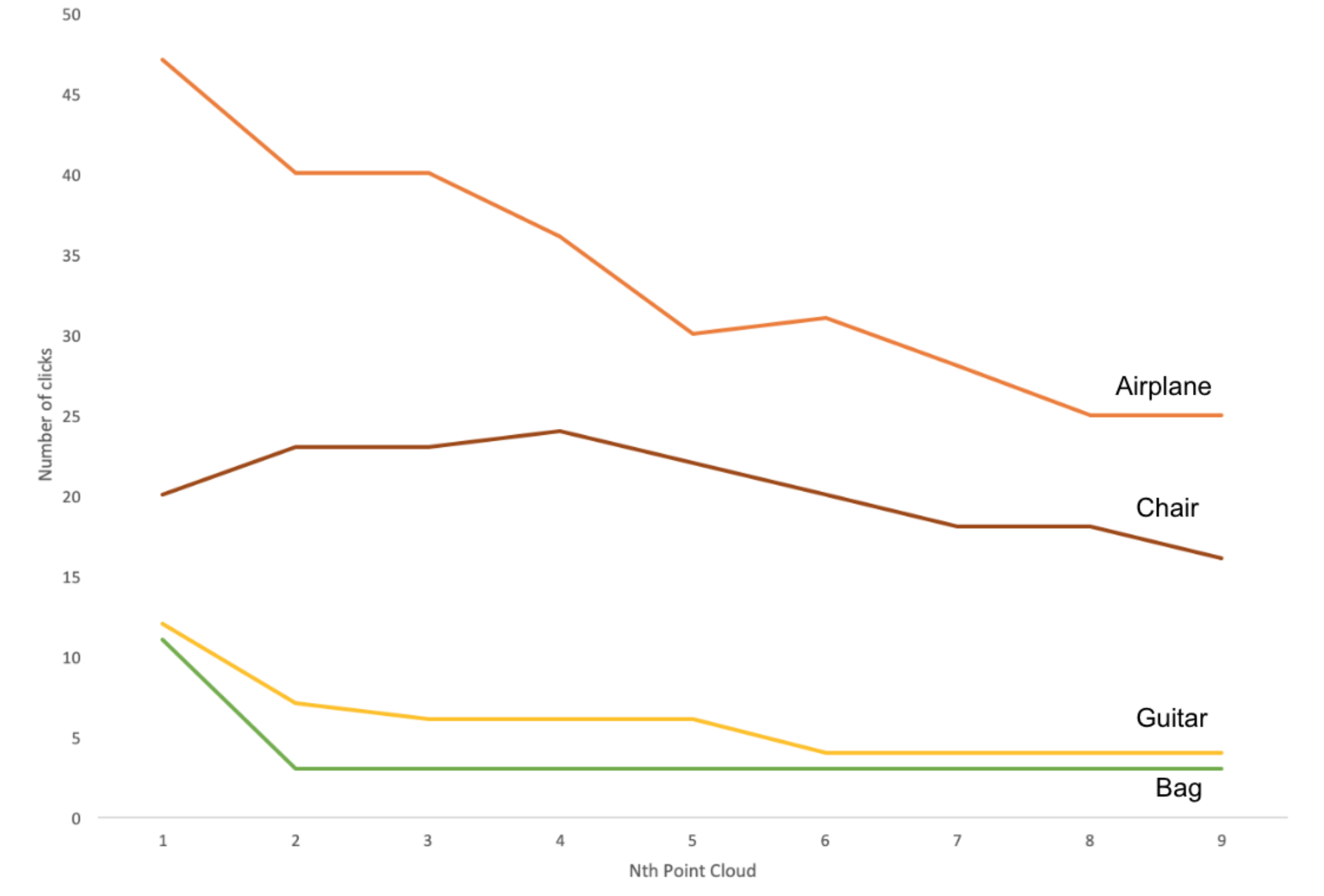}
        \caption{
                \label{fig:MulitplePointClouds} % 
                Number of clicks taken to annotate subsequent point clouds using our framework. We see a reduction in number of clicks needed as more point clouds of the new dataset are annotated.
        }
\end{figure}

With subsequent complete annotations of point clouds coming from the same dataset, we expect a reduction in the human supervision needed in order to have a scalable system. As we incrementally train the network on a progressively complete annotation of the point clouds, the model adapts to the properties of the new domain represented by the dataset. Thus, we are able to predict a more accurate segmentation of the point cloud in the initial iterations thereby cutting down on the total number of user correction steps needed. This is validated via our experiments as illustrated in Figure \ref{fig:MulitplePointClouds}. 

While previous works measure the amount of user supervision based on invested time, we focused on quantifying supervision via number of clicks. Through our experiments we observed that time taken to annotate a point clouds reduces with the number of point clouds annotated even in completely manual methods. This is because a large part of the time taken in annotating goes in manipulating the point clouds on a 2D tool. As the annotators label more point clouds, they get more accustomed to the tool and the relevant manipulation interactions, reducing the overall time they need in annotating subsequent point clouds. On the other hand, the number of clicks needed depend more on complexity of the point cloud and the number of classes to be annotated instead of the number of previous point clouds annotated in the system making it a suitable metric for evaluation.

\section{Conclusion}
We provide a scalable interactive learning framework that can be used to annotate large point cloud datasets. By fusing together three different cues (human annotations, learnt semantic similarity and geometric consistencies) we are able to obtain accurate annotations with fewer human interactions. We note that while the number of clicks are a useful proxy for the quantum of human interaction needed, it is also important to study the amount of time needed for each click as it adds to overall human time investment needed in annotating a dataset. Significant leaps in reducing the cognitive overload for a human annotator can be made by replacing 2D user interfaces with spatial user interfaces facilitated via virtual reality systems as they make point cloud manipulation and visualization more natural of the annotator.     

\bibliography{pc_segmentation}

\begin{thebibliography}{16}
\providecommand{\natexlab}[1]{#1}
\providecommand{\url}[1]{\texttt{#1}}
\expandafter\ifx\csname urlstyle\endcsname\relax
  \providecommand{\doi}[1]{doi: #1}\else
  \providecommand{\doi}{doi: \begingroup \urlstyle{rm}\Url}\fi

\bibitem[Behley et~al.(2019)Behley, Garbade, Milioto, Quenzel, Behnke,
  Stachniss, and Gall]{Behley2019ADF}
Behley, J., Garbade, M., Milioto, A., Quenzel, J., Behnke, S., Stachniss, C.,
  and Gall, J.
\newblock A dataset for semantic segmentation of point cloud sequences.
\newblock \emph{CoRR}, abs/1904.01416, 2019.

\bibitem[Benhabiles et~al.(2009)Benhabiles, Vandeborre, Lavou{\'{e}}, and
  Daoudi]{segmentation_eval}
Benhabiles, H., Vandeborre, J., Lavou{\'{e}}, G., and Daoudi, M.
\newblock A framework for the objective evaluation of segmentation algorithms
  using a ground-truth of human segmented 3d-models.
\newblock In \emph{{IEEE} International Conference on Shape Modeling and
  Applications, {SMI} 2009, Beijing, China, 26-28 June 2009}, pp.\  36--43,
  2009.
\newblock \doi{10.1109/SMI.2009.5170161}.
\newblock URL \url{https://doi.org/10.1109/SMI.2009.5170161}.

\bibitem[Chang et~al.(2015{\natexlab{a}})Chang, Funkhouser, Guibas, Hanrahan,
  Huang, Li, Savarese, Savva, Song, Su, Xiao, Yi, and Yu]{shapenet2015}
Chang, A.~X., Funkhouser, T., Guibas, L., Hanrahan, P., Huang, Q., Li, Z.,
  Savarese, S., Savva, M., Song, S., Su, H., Xiao, J., Yi, L., and Yu, F.
\newblock {ShapeNet: An Information-Rich 3D Model Repository}.
\newblock Technical Report arXiv:1512.03012 [cs.GR], Stanford University ---
  Princeton University --- Toyota Technological Institute at Chicago,
  2015{\natexlab{a}}.

\bibitem[Chang et~al.(2015{\natexlab{b}})Chang, Funkhouser, Guibas, Hanrahan,
  Huang, Li, Savarese, Savva, Song, Su, Xiao, Yi, and
  Yu]{DBLP:journals/corr/ChangFGHHLSSSSX15}
Chang, A.~X., Funkhouser, T.~A., Guibas, L.~J., Hanrahan, P., Huang, Q., Li,
  Z., Savarese, S., Savva, M., Song, S., Su, H., Xiao, J., Yi, L., and Yu, F.
\newblock Shapenet: An information-rich 3d model repository.
\newblock \emph{CoRR}, abs/1512.03012, 2015{\natexlab{b}}.
\newblock URL \url{http://arxiv.org/abs/1512.03012}.

\bibitem[Choi et~al.(2016)Choi, Zhou, Miller, and Koltun]{Vladlen}
Choi, S., Zhou, Q.-Y., Miller, S., and Koltun, V.
\newblock A large dataset of object scans.
\newblock \emph{arXiv:1602.02481}, 2016.

\bibitem[Hackel et~al.(2017)Hackel, Savinov, Ladicky, Wegner, Schindler, and
  Pollefeys]{hackel2017isprs}
Hackel, T., Savinov, N., Ladicky, L., Wegner, J.~D., Schindler, K., and
  Pollefeys, M.
\newblock {SEMANTIC3D.NET: A new large-scale point cloud classification
  benchmark}.
\newblock In \emph{ISPRS Annals of the Photogrammetry, Remote Sensing and
  Spatial Information Sciences}, volume IV-1-W1, pp.\  91--98, 2017.

\bibitem[{Keselman} et~al.(2017){Keselman}, {Woodfill}, {Grunnet-Jepsen}, and
  {Bhowmik}]{8014901}
{Keselman}, L., {Woodfill}, J.~I., {Grunnet-Jepsen}, A., and {Bhowmik}, A.
\newblock Intel(r) realsense(tm) stereoscopic depth cameras.
\newblock In \emph{2017 IEEE Conference on Computer Vision and Pattern
  Recognition Workshops (CVPRW)}, pp.\  1267--1276, July 2017.
\newblock \doi{10.1109/CVPRW.2017.167}.

\bibitem[Landrieu \& Simonovsky(2017)Landrieu and
  Simonovsky]{DBLP:journals/corr/abs-1711-09869}
Landrieu, L. and Simonovsky, M.
\newblock Large-scale point cloud semantic segmentation with superpoint graphs.
\newblock \emph{CoRR}, abs/1711.09869, 2017.
\newblock URL \url{http://arxiv.org/abs/1711.09869}.

\bibitem[Mishra(1997)]{nearest_neighbor}
Mishra, B.
\newblock \emph{Computational real algebraic geometry}, pp.\  537--558.
\newblock CRC Press, 7 1997.

\bibitem[Mo et~al.(2018)Mo, Zhu, Chang, Yi, Tripathi, Guibas, and
  Su]{DBLP:journals/corr/abs-1812-02713}
Mo, K., Zhu, S., Chang, A.~X., Yi, L., Tripathi, S., Guibas, L.~J., and Su, H.
\newblock Partnet: {A} large-scale benchmark for fine-grained and hierarchical
  part-level 3d object understanding.
\newblock \emph{CoRR}, abs/1812.02713, 2018.
\newblock URL \url{http://arxiv.org/abs/1812.02713}.

\bibitem[{Newcombe} et~al.(2011){Newcombe}, {Izadi}, {Hilliges}, {Molyneaux},
  {Kim}, {Davison}, {Kohi}, {Shotton}, {Hodges}, and {Fitzgibbon}]{Newcombe}
{Newcombe}, R.~A., {Izadi}, S., {Hilliges}, O., {Molyneaux}, D., {Kim}, D.,
  {Davison}, A.~J., {Kohi}, P., {Shotton}, J., {Hodges}, S., and {Fitzgibbon},
  A.
\newblock Kinectfusion: Real-time dense surface mapping and tracking.
\newblock In \emph{2011 10th IEEE International Symposium on Mixed and
  Augmented Reality}, pp.\  127--136, Oct 2011.
\newblock \doi{10.1109/ISMAR.2011.6092378}.

\bibitem[Qi et~al.(2016)Qi, Su, Mo, and Guibas]{pointnet}
Qi, C.~R., Su, H., Mo, K., and Guibas, L.~J.
\newblock Pointnet: Deep learning on point sets for 3d classification and
  segmentation.
\newblock \emph{CoRR}, abs/1612.00593, 2016.
\newblock URL \url{http://arxiv.org/abs/1612.00593}.

\bibitem[Roynard et~al.(2018)Roynard, Deschaud, and
  Goulette]{Roynard2018ParisLille3DAL}
Roynard, X., Deschaud, J.-E., and Goulette, F.
\newblock Paris-lille-3d: a large and high-quality ground truth urban point
  cloud dataset for automatic segmentation and classification.
\newblock \emph{I. J. Robotics Res.}, 37:\penalty0 545--557, 2018.

\bibitem[Tchapmi et~al.(2017)Tchapmi, Choy, Armeni, Gwak, and
  Savarese]{DBLP:journals/corr/abs-1710-07563}
Tchapmi, L.~P., Choy, C.~B., Armeni, I., Gwak, J., and Savarese, S.
\newblock Segcloud: Semantic segmentation of 3d point clouds.
\newblock \emph{CoRR}, abs/1710.07563, 2017.
\newblock URL \url{http://arxiv.org/abs/1710.07563}.

\bibitem[Yi et~al.(2016)Yi, Kim, Ceylan, Shen, Yan, Su, Lu, Huang, Sheffer, and
  Guibas]{Yi:2016:SAF:2980179.2980238}
Yi, L., Kim, V.~G., Ceylan, D., Shen, I.-C., Yan, M., Su, H., Lu, C., Huang,
  Q., Sheffer, A., and Guibas, L.
\newblock A scalable active framework for region annotation in 3d shape
  collections.
\newblock \emph{ACM Trans. Graph.}, 35\penalty0 (6):\penalty0 210:1--210:12,
  November 2016.
\newblock ISSN 0730-0301.
\newblock \doi{10.1145/2980179.2980238}.
\newblock URL \url{http://doi.acm.org/10.1145/2980179.2980238}.

\bibitem[Zhang(2012)]{Zhang:2012:MKS:2225053.2225203}
Zhang, Z.
\newblock Microsoft kinect sensor and its effect.
\newblock \emph{IEEE MultiMedia}, 19\penalty0 (2):\penalty0 4--10, April 2012.
\newblock ISSN 1070-986X.
\newblock \doi{10.1109/MMUL.2012.24}.
\newblock URL \url{http://dx.doi.org/10.1109/MMUL.2012.24}.

\end{thebibliography}
\bibliographystyle{icml2019}

\end{document}